\pdfoutput=1

\documentclass[11pt]{article}

\usepackage[final]{acl}

\usepackage{times}
\usepackage{latexsym}

\usepackage[T1]{fontenc}

\usepackage[utf8]{inputenc}

\usepackage{microtype}

\usepackage{inconsolata}

\usepackage{graphicx}

\usepackage[utf8]{inputenc} 
\usepackage[T1]{fontenc}    
\usepackage{hyperref}       
\usepackage{url}            
\usepackage{booktabs}       
\usepackage{amsfonts}       
\usepackage{nicefrac}       
\usepackage{microtype}      
\usepackage{xcolor}         
\usepackage{subfigure}
\usepackage{amsmath}
\usepackage{amssymb}
\usepackage{mathtools}
\usepackage{amsthm}
\usepackage{multicol}
\usepackage{multirow}
\usepackage{tablefootnote}
\usepackage{pifont}
\usepackage{makecell}
\theoremstyle{plain}

\newtheorem{theorem}{Theorem}[section]

\theoremstyle{definition}
\newtheorem{definition}[theorem]{Definition}

\theoremstyle{remark}

\usepackage[ruled,longend,noline]{algorithm2e}

\SetCommentSty{mycommfont}

\newcommand{\name}{Ouroboros}
\title{\name: Generating Longer Drafts Phrase by Phrase\\ for Faster Speculative Decoding}

\author{
 \textbf{Weilin Zhao\textsuperscript{1}\thanks{\ \ indicates equal contribution.}},
 \textbf{Yuxiang Huang\textsuperscript{1}$^*$},
 \textbf{Xu Han\textsuperscript{1,2}\thanks{\ \ indicates corresponding authors.}},
 \textbf{Wang Xu\textsuperscript{1}},
 \textbf{Chaojun Xiao\textsuperscript{1}},
\\
 \textbf{Xinrong Zhang\textsuperscript{1}},
 \textbf{Yewei Fang\textsuperscript{3}},
 \textbf{Kaihuo Zhang\textsuperscript{3}},
 \textbf{Zhiyuan Liu\textsuperscript{1}$^\dag$},
 \textbf{Maosong Sun\textsuperscript{1}}
\\
\\
 \textsuperscript{1}{Department of Computer Science and Technology, Institute for Artificial Intelligence,}
 \\
 \hspace{0.4em}\textsuperscript{}{Beijing National Research Center for Information Science and Technology,}
 \\
 \hspace{0.4em}\textsuperscript{}{Tsinghua University, Beijing, China.}
 \\
\textsuperscript{2}{Shanghai Artificial Intelligence Laboratory, Shanghai, China.}
 \textsuperscript{3}{ModelBest Inc.}
\\
{\tt \{zwl23,huang-yx21\}@mails.tsinghua.edu.cn, \{hanxu2022,liuzy\}@tsinghua.edu.cn}
}

\begin{document}
\maketitle
\begin{abstract}
Speculative decoding is a widely used method that accelerates the generation process of large language models (LLMs) with no compromise in model performance.
It achieves this goal by using an existing smaller model for drafting and then employing the target LLM to verify the draft in a low-cost parallel manner.
Under such a drafting-verification framework, drafting efficiency has become a bottleneck in the final speedup of speculative decoding. Therefore, generating longer drafts at less cost can lead to better decoding speedup.
To achieve this, we introduce \name, which can generate draft phrases to parallelize the drafting process and meanwhile lengthen drafts in a training-free manner.
The experimental results on various typical text generation tasks show that \name~can achieve speedups of up to $2.8\times$ over speculative decoding and $3.9\times$ over vanilla decoding, without fine-tuning draft and target models.
The source code of \name~is available at \href{https://github.com/thunlp/Ouroboros}{https://github.com/thunlp/Ouroboros}.
\end{abstract}

\section{Introduction}

Benefiting from recent advances in parallel computing devices and distributed training algorithms~\cite{megatron,deepspeed}, the training time of LLMs has been significantly shortened.
However, it is still challenging to achieve parallelization in model inference because most LLMs rely on autoregressive generation mechanisms that exhibit sequential dependencies among generated tokens. For these autoregressive LLMs, they have to generate tokens one by one.

To accelerate the inference of LLMs, typical model compression methods such as quantization~\cite{frantar2023gptq,awq,squeezellm} and pruning~\cite{Han2015LearningBW, wang-etal-2020-structured,xia2023sheared} may cause model performance degradation and sometimes even require non-negligible additional training costs.
To losslessly accelerate the inference of LLMs, speculative decoding~\cite{microsoftspec,googlespec,deepmindspec} has been proposed.
Given a target LLM, speculative decoding selects a much smaller LLM as the draft model to generate multiple tokens as a draft. The target LLM then reviews the draft in parallel, accepting the longest accurate prefix while discarding the remaining draft tokens.

\begin{figure}[t]
  \centering
    \includegraphics[width=0.5\textwidth]{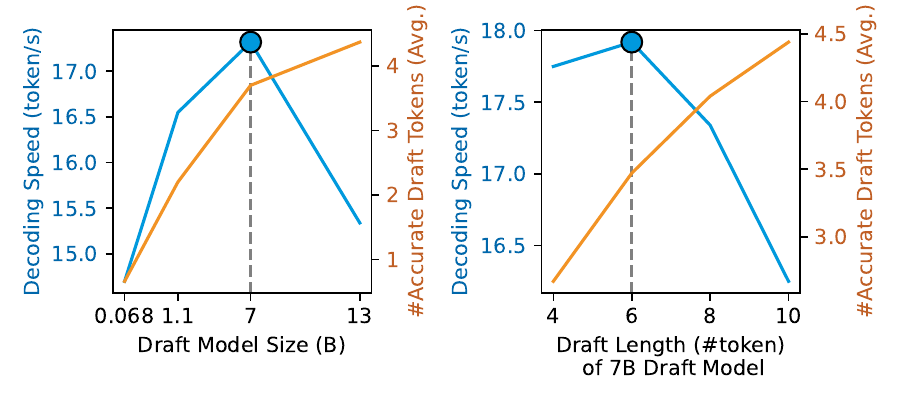}
    \caption{The trade-off between drafting efficiency and effectiveness. The figure illustrates the optimal speculative decoding speed of Llama-2-chat-70B on MT-Bench~\cite{zheng2024judging} and the corresponding average number of accurate draft tokens at each iteration.}
  \label{fig:tradeoff}
\end{figure}

Under such a drafting-verification framework, how to efficiently generate long and accurate drafts has become a critical factor in accelerating LLM inference.
As in Figure~\ref{fig:tradeoff}, we examine the trade-off between draft model size and speculative decoding speed. 
We observe that larger draft models tend to achieve higher draft accuracy, but the cost of generating drafts also rises, so that medium-sized models offer the best decoding speed. Besides, we investigate the trade-off between draft length and speculative decoding speed. Our findings indicate that the target model can accept more tokens per iteration on average when the draft is longer, but longer drafts
introduce more forward passes required for the draft model, and thus, the optimal draft length in speculative decoding is not the largest one.


Based on the further analysis of the above pilot experimental results, it is not difficult to observe the limitations of speculative decoding regarding drafting efficiency:
\textbf{(1) Insufficient Draft}. Drafting too short misses potential acceleration. However, because of the time overhead involved in drafting itself, generating long drafts may result in high costs when these drafts fail. If longer drafts can be generated more efficiently, the acceleration effect will be enhanced significantly.
\textbf{(2) Underutilized Draft}. For current speculative decoding, tokens that are not accepted by the target model are completely discarded, resulting in a high failure cost when generating long drafts. In fact, some of the discarded tokens may contain useful information that could be utilized in future drafting iterations.

To overcome the above limitations, this paper introduces a more efficient decoding framework named \name, whose improvements over speculative decoding are as follows:

\textbf{(1) Accelerating drafting via phrases.} Given that model generation is memory-bound rather than computation-bound~\cite{googlespec}, drafting at the phrase level rather than the token level can make the drafting phase more efficient at producing longer drafts. 
Inspired by previous efforts such as lookahead decoding~\cite{fu2023lookahead} that successfully employed phrases to accelerate the target model, \name~adapts this method to enhance drafting efficiency. In subsequent sections, we will demonstrate that accelerating the draft model with phrases first and then using the draft to accelerate the target model will be more efficient than directly accelerating the target model with phrases.

\textbf{(2) Lengthening drafts via phrases.} Concatenating phrases can extend the draft even longer at a low cost. According to the last token of the draft, phrases starting with the last token can be used to extend the draft. By concatenating the draft and  $k$ different phrases, \name~generates $k$ longer drafts for the target model to verify, introducing almost zero additional costs.

\textbf{(3) Generating phrases from verification.} During the verification phase, \name~filters out high-quality phrases from those discarded tokens in speculative decoding. These discarded phrases can be used to accelerate drafting in subsequent iterations.

\textbf{(4) Reusing phrases from history contexts.} We find that phrases generated from similar tasks can be reused to speed up drafting with each other. \name~thus reuses phrases in history contexts to accelerate drafting in subsequent iterations.

\textbf{Notably, \name{} does not require any additional training, and can be applied in all applications with speculative decoding.}
We implement \name~and conduct sufficient experiments on various text generation tasks such as code generation~\cite{humaneval,mbpp}, text summarization~\cite{cnndm1,cnndm2} and machine translation~\cite{wmt16}.
The experimental results demonstrate that \name~is completely lossless on task performance and can achieve significant inference acceleration without additional model fine-tuning. Compared with the recent competitive decoding methods, \name~achieves speedups of up to 1.9$\times$ over lookahead decoding, up to 2.8$\times$ over speculative decoding, and up to 3.9$\times$ over naive autoregressive decoding.

\begin{figure*}[t]
  \centering
    \includegraphics[width=0.97\textwidth]{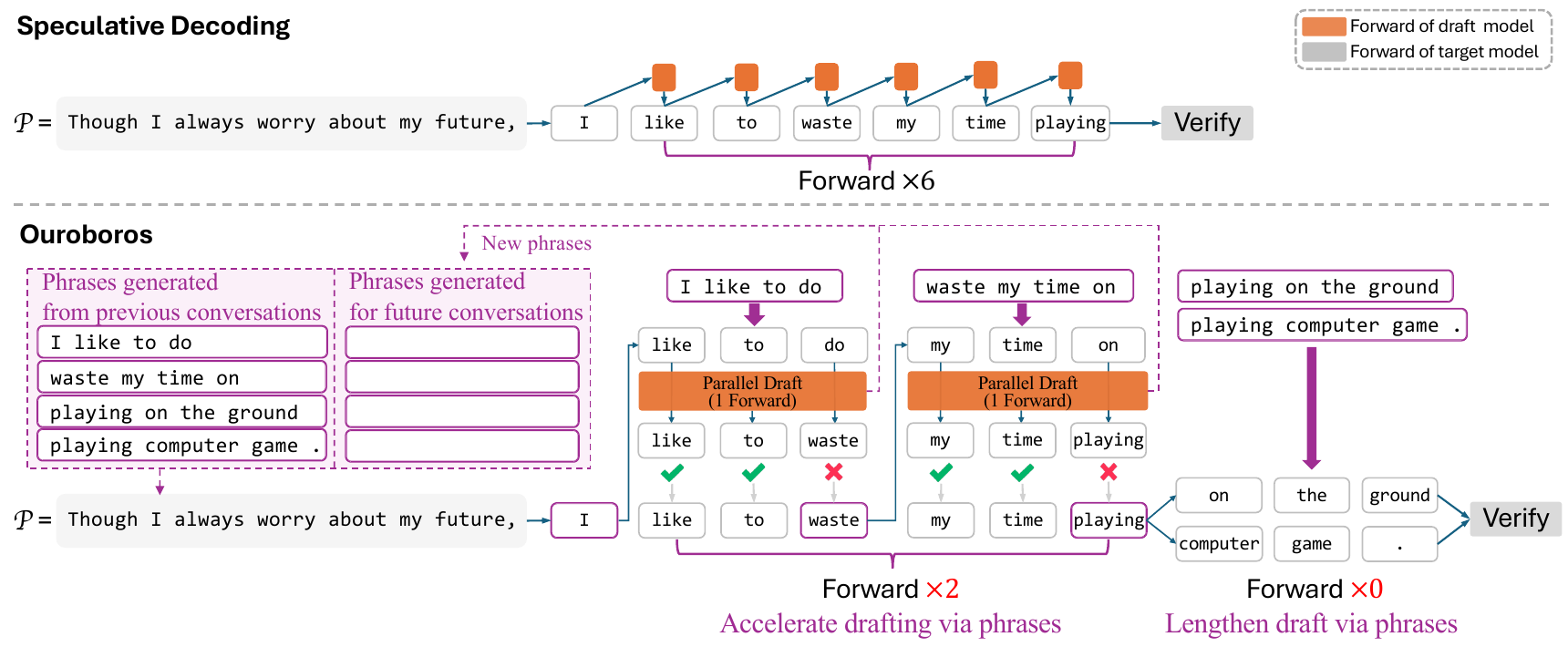}
    \caption{The framework of \name, which achieves better drafting efficiency than vanilla speculative decoding while also lengthening the drafts.}
  \label{fig:framework}
\end{figure*}

\section{Preliminary}
\label{sec:preliminary}

For a target model $\mathcal{T}$ to accelerate inference, speculative decoding first finds a suitable draft model $\mathcal{S}$.
Given an input prefix $x_{1\cdots n}$, speculative decoding uses the draft model $\mathcal{S}$ to generate a draft $d_{1\cdots \gamma}$ consisting of $\gamma$ tokens in an autoregressive manner,
\begin{equation}
\small
\begin{aligned}
d_1 &= \arg \max_{d_1} P_{\mathcal{S}}(d_1 ~|~ x_{1\cdots n}),\\
d_i &= \arg \max_{d_i} P_{\mathcal{S}}(d_i ~|~ x_{1\cdots n}, d_{1\cdots i-1}),~i=2\cdots \gamma.
\end{aligned}
\end{equation}
Then, the target model $\mathcal{T}$ verifies the draft $d_{1\cdots \gamma}$, by inputting $d_{1\cdots \gamma}$ and conducting the next token prediction task using a single forward. Each verification result $v_i$ is sampled from the next token prediction distribution of the target model,
\begin{equation}
\small
\begin{aligned}
v_1 &\sim P_{\mathcal{T}}(\cdot ~|~ x_{1\cdots n}),\\
v_i &\sim P_{\mathcal{T}}(\cdot ~|~ x_{1\cdots n}, d_{1\cdots i-1}),~i=2\cdots \gamma.
\end{aligned}
\end{equation}
There must exist an $A\in[0, \gamma]$ satisfying that
\begin{equation}
\small
\left\{
\begin{aligned}
v_{1\cdots A} &= d_{1\cdots A}, \\
v_{A+1} &\neq d_{A+1} ~\text{or}~ A=\gamma.
\end{aligned}
\right.
\label{eq:A_def}
\end{equation}
Here, $v_{1\cdots A+1}$ will be the same as the result when the target model autoregressively generates $A+1$ tokens from $x$, because at this time,
\begin{equation}
\small
\begin{aligned}
v_1 &\sim P_{\mathcal{T}}(\cdot ~|~ x_{1\cdots n}),\\
v_i &\sim P_{\mathcal{T}}(\cdot ~|~ x_{1\cdots n}, v_{1\cdots i-1}),~i=2\cdots A+1.
\end{aligned}
\end{equation}

We call $A$ the number of accurate tokens in the draft, $A+1$ tokens $d_{1\cdots A}, v_{A+1}$ are accepted as the generation result of the target model while the remaining draft tokens $d_{A+1 \cdots \gamma}$ are discarded.
Please note that the derivation above differs slightly from the original speculative decoding~\cite{deepmindspec}: we set the temperature for the random sampling in the draft model to 0 for a clearer explanation, but this would not affect the correctness of the target model.

We define $A(\gamma)$ as the average number of accurate tokens when the draft model is $\mathcal{S}$ and the draft length is $\gamma$.
The speedup of speculative decoding is calculated as follows~\cite{googlespec}, where $t_{\mathcal{S}}$ and $t_{\mathcal{T}}$ are the forward time of the draft model and the target model, respectively,
\begin{equation}
\small
{[A(\gamma)+1] \cdot t_{\mathcal{T}} \over \gamma \cdot t_{\mathcal{S}} + t_{\mathcal{T}}},
\label{eq:speedup}
\end{equation}
where the numerator is the time for the vanilla generation of the target model, and the denominator is the time of the speculative decoding.


\section{Method}

In \name, we first accelerate drafting via phrases, so that we can generate $\gamma$ tokens with fewer forward passes of the draft model, and we denote this reduction ratio as $c$.
Subsequently, due to the fact that function $A(\gamma)$ is monotonically non-decreasing, we lengthen drafts via phrases, enabling the generation of more tokens without the need for additional forward passes. We denote this costless extended length as $\beta$.
Following these two optimization directions, Eq.~(\ref{eq:speedup}) becomes
\begin{equation}
\small
{[A(\gamma + \beta)+1] \cdot t_{\mathcal{T}} \over {1\over c}\gamma \cdot t_{\mathcal{S}} + t_{\mathcal{T}}}.
\label{eq:speedup_our}
\end{equation}
We further enlarge $c$ by heuristically generating phrases from verification results and reusing phrases from historically generated contexts, which can lead to better drafting efficiency. Next, we will delve into the details for each part of \name.

\subsection{Accelerating Drafting via Phrases}

We get inspiration from lookahead decoding~\cite{fu2023lookahead}, which uses phrases to directly accelerate the target model $\mathcal{T}$. However, each round of phrase drafting requires a forward pass of the target model $\mathcal{T}$ to verify the draft, limiting the whole acceleration effect of lookahead decoding. 
Different from lookahead decoding, we use phrases to indirectly accelerate the target model $\mathcal{T}$ through a draft model $\mathcal{S}$, allowing each forward pass of the target model to simultaneously verify multiple rounds of phrases, achieving a better acceleration.
As shown in Figure~\ref{fig:framework}, in \name, the drafting process of the draft model is performed phrase by phrase instead of token by token. Multiple new phrases are generated in parallel during each forward of the draft model. 
We will not go into detail how to generate new draft phrases in parallel due to space constraints, and can refer to \citet{fu2023lookahead}.

\subsection{Lengthening Drafts via Phrases}
\label{sec:lengthen}

Given that model generation is memory-bound, the time it takes for the target model to verify dozens of tokens using a single forward is not much different from the time spent on verifying a single token~\cite{googlespec}. Therefore, we propose to use phrases to heuristically extend drafts since phrase concatenation is nearly zero-cost.

\begin{figure}[t]
  \centering
    \includegraphics[width=0.7\linewidth]{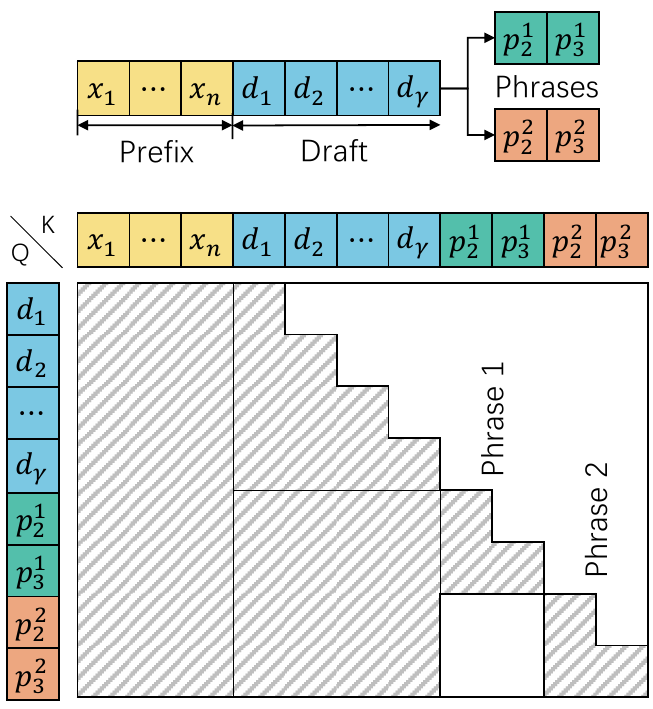}
  \caption{The customized attention masking mechanism for verifying lengthened drafts.}
  \label{fig:attn_mask}
\end{figure}

Trying multiple phrases to get multiple longer drafts can increase the probability that one of these phrases passes the verification of the target model. But the cost of verifying these longer drafts one by one is also unbearable. Inspired by~\citet{medusa}, we construct a sophisticated Transformer attention masking mechanism to complete the verification of all longer drafts with only one forward pass of the target model $\mathcal{T}$, as shown in Figure~\ref{fig:attn_mask}.

Specifically, given a draft $d_{1\cdots\gamma}$, we select out $K$ phrases $p^1_{1\cdots \beta}, \cdots, p^K_{1\cdots \beta}$ starting with the token $d_{\gamma}$, i.e., $p^1_1=p^2_1=\cdots=p^K_1=d_\gamma$. The $K$ lengthened drafts are
\begin{equation}
\label{eq:draft-process}
\small
\left\{
\begin{aligned}
d_{1\cdots \gamma}&, p^1_{2\cdots \beta},\\
d_{1\cdots \gamma}&, p^2_{2\cdots \beta},\\
\vdots\\
d_{1\cdots \gamma}&, p^K_{2\cdots \beta}.\\
\end{aligned}
\right.
\end{equation}
The target model then calculates $v^j_i$ to verify $p^j_i$ for all $j=2\cdots K, i=2\cdots \beta+1$ in Eq.~(\ref{eq:verify_suffix}). All $v^j_i$ are computed in parallel by a single forward using the attention masking mechanism in Figure~\ref{fig:attn_mask}.
\begin{equation}
\small
v^j_i \sim P_{\mathcal{T}}(\cdot ~|~ x_{1\cdots n}, d_{1\cdots \gamma}, p^j_{2\cdots i-1})
\label{eq:verify_suffix}
\end{equation}

Similar to Eq.~(\ref{eq:A_def}), we define $\hat A^j$ satisfying
\begin{equation}
\small
\left\{
\begin{aligned}
v^j_{2\cdots \hat A^j} &= p^j_{2\cdots \hat A^j}, \\
v^j_{\hat A^j+1} &\neq p^j_{\hat A^j+1} ~\text{or}~ \hat A^j=\beta.
\end{aligned}
\right.
\label{eq:hatA_def}
\end{equation}
Once the draft $d_{1\cdots \gamma}$ is fully accepted by the target model, we use the phrase $p^j$ with the largest $\hat A^j$ to enlarge the number of accepted tokens. More specifically, $d_{1\cdots \gamma}, p^j_{2\cdots \hat A^j}, v^j_{\hat A^j+1}$ will be accepted.

\begin{figure*}[t]
  \centering
    \includegraphics[width=1\textwidth]{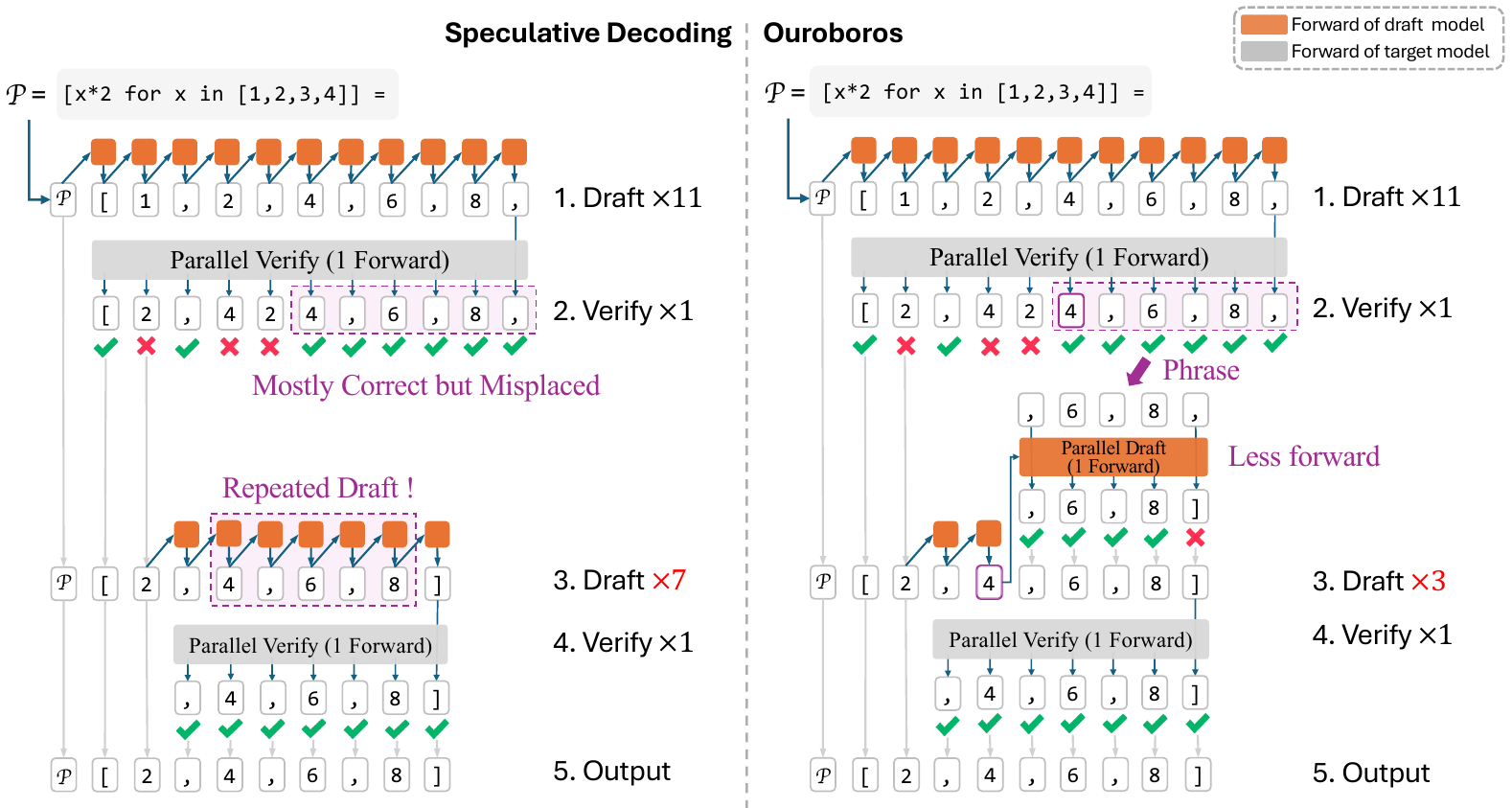}
    \caption{The illustration of generating phrases from the verification results of the target model.}
  \label{fig:inspiration}
\end{figure*}

\begin{table}[t]
\center
\scalebox{0.85}{
\begin{tabular}{c|c|c|c}
\toprule
Yi-base 34B/6B & MBPP & CNN/DM & WMT-16 \\
\midrule
$A$ & 12.7 & 8.4 & 5.3 \\
\#Match & 18.9 & 17.8 & 17.0 \\
\bottomrule
\end{tabular}
}
\caption{Average matched tokens compared with the number of accepted tokens. Experiment on Yi-base 34B/6B~\cite{Yi} on three datasets, MBPP~\cite{mbpp}, CNN/DM~\cite{cnndm1, cnndm2} and WMT16~\cite{wmt16} using Speculative Decoding, with draft length $\gamma$ = 20.}
\label{tab:accept_vs_match}
\end{table}

\subsection{Generating Phrases from Verification}

In speculative decoding, the cost when a draft fails is relatively high.
If draft $d_{1\cdots \gamma}$ is not fully accepted, traditional verification only accepts the correct draft prefix $d_{1\cdots A}$ and discards $d_{A+1\cdots \gamma}$, 
as in Section~\ref{sec:preliminary}.
To evaluate the information in these discarded draft tokens, we define \#Match$(d_{1\cdots \gamma}, v_{1\cdots \gamma})$ the number of the matched tokens of the draft and the verification result,
\begin{equation}
\small
\text{\#Match}(d_{1\cdots \gamma}, v_{1\cdots \gamma}) = \sum_{i=1}^{\gamma} [d_i = v_i],
\end{equation}
where $[d_i = v_i] = 1$ when $d_i = v_i$ and $0$ otherwise.

As in Table~\ref{tab:accept_vs_match}, we find that \#Match is much larger than the length of the accurate prefix $A$. We view some specific cases and find that sometimes the emergence of this situation is caused by the misplacement of the generation, as shown in Figure~\ref{fig:inspiration}.
Therefore, we propose to filter out those sub-segments of the discarded draft that match the verification results of the target model, as shown in Figure~\ref{fig:inspiration}. We then insert those segments into the phrases pool to accelerate future drafts.

As in Section~\ref{sec:lengthen}, if draft $d_{1\cdots \gamma}$ is fully accepted, only one phrase suffix $p^j$ is used, leaving all other tried phrases unused. We suppose the verification result $v^o_{2\cdots \beta}$ fixes some errors in the phrase $p^o_{2\cdots \beta}$ for all $o\neq j$: we use $p^o_1, v^o_{2\cdots \beta}$ to replace the phrase $p^o_{1\cdots \beta}$ in the phrase pool.


\subsection{Reusing Phrases from History Contexts}

In real-world scenarios, the adjacent conversations from users may exhibit similarities. Leveraging these similarities, we can further enhance the speed of the model inference.
Different from lookahead decoding that cleanup phrases from history contexts when conducting the generation of subsequent input queries, we propose to reuse phrases generated from previous generations to further accelerate the drafting process.

\subsection{The Advantages
 of the Training-free Framework}

\name~is entirely training-free.
We have not employed methods like model distillation to increase the function $A(\gamma)$ or model compression to decrease $t_S$ in Eq.~(\ref{eq:speedup}).
In terms of phrase generation, all our phrases generation strategies are heuristic. All phrases are gradually accumulated during the generation process of models, without prior preparation on a large-scale corpus.
All these mean that, given a draft model in any speculative decoding method, we can use \name~to help these methods achieve further speedup without introducing additional costs.

\section{Experiments}

This section focuses on evaluating \name~on various text generation tasks to demonstrate the efficiency and effectiveness of \name.

\subsection{Experimental Settings}

In order to evaluate the overall acceleration caused by \name, we evaluate \name~on various typical text generation tasks, including code generation, arithmetic reasoning, document summarization, and machine translation.

\textbf{Datasets}. For code generation, we evaluate \name~on HumanEval~\cite{humaneval} and the validation set of MBPP~\cite{mbpp}. There are 164 entries in HumanEval and they are composed of a text prompt and a prefix of Python function. The validation set of MBPP has 90 entries, in which the whole function is expected to be predicted with a given text prompt and test cases. The maximum generation lengths on HumanEval and MBPP are set to 512.
For arithmetic reasoning, document summarization and machine translation, we evaluate our method on GSM8K~\cite{gsm8k}, CNN/DM~\cite{cnndm1, cnndm2} and WMT16~\cite{wmt16}, respectively. We randomly sample 100 entries from GSM8K and CNN/DM, and sample 100 entries from the German-to-English translation subset of WMT16. The maximum generation lengths on GSM8k, CNN/DM and WMT16 are respectively set to 256, 128, and 64.

\begin{figure*}[t]
  \centering
    \includegraphics[width=0.9\textwidth]{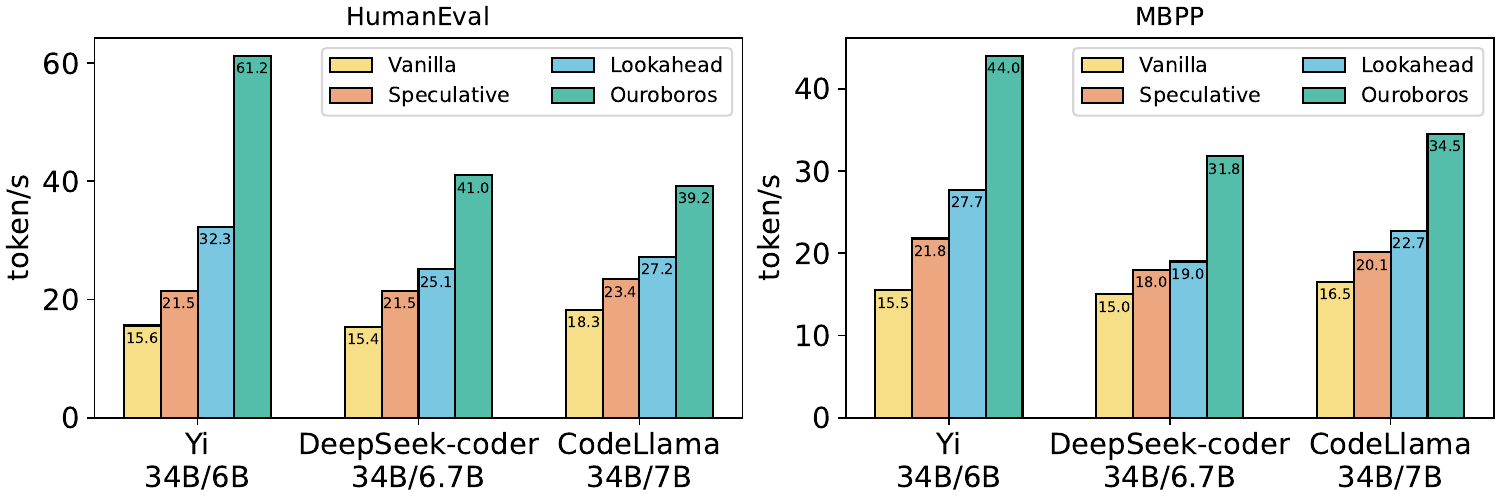}
  \caption{The greedy decoding speed (token/s) on HumanEval and MBPP.}
  \label{fig:greedy_code}
\end{figure*}

\textbf{Models}. For HumanEval and MBPP, we use Yi-base-34B/6B~\cite{Yi}, DeepSeek-coder-instruct-33b/6.7B~\cite{bi2024deepseek} and CodeLlama-instruct-34B/7B~\cite{roziere2023code} as the target/draft models for our experiments. For GSM8K, CNN/DM and WMT16, we use Yi-base-34B/6B and Llama-2-chat-70B/7B~\cite{llama2}. We use the larger model as the target model and the smaller one for drafting. All these models are recently representative and popular LLMs.

\textbf{Evaluation Methods}. The baselines of our experiments are vanilla autoregressive decoding, speculative decoding~\cite{googlespec, deepmindspec}, and lookahead decoding~\cite{fu2023lookahead}. We report the decoding speed (token/s) and the speedup ratio compared with vanilla autoregressive decoding. More hyperparameters of our experiments are included in Appendix \ref{sec:hyperparameters}.

\textbf{Hardware and Implementation}. All experiments are performed on 2 $\times$ NVIDIA 80GB A800 GPU with NVLINK $\times$ 8 interconnected. The CPU used is the Intel(R) Xeon(R) Platinum 8350C. We use the Huggingface transformers package to conduct automatic model parallelism for our methods and all baselines.

\subsection{Overall Results}

\bgroup
\small
\setlength{\tabcolsep}{4pt}
\begin{table}[t]
\center
\scalebox{0.76}{
\begin{tabular}{ll|cc|cc}
\toprule
\multicolumn{1}{l}{\multirow{2}{*}{Task}}&\multicolumn{1}{l|}{\multirow{2}{*}{Algorithm}} & \multicolumn{2}{c|}{\makecell{Yi 34B/6B}} & \multicolumn{2}{c}{\makecell{Llama-2 70B/7B}} \\
& & token/s & speedup & token/s & speedup \\
\midrule
\multirow{4}{*}{GSM8k} & Vanilla & 15.33 & 1.00$\times$ & 8.96 & 1.00$\times$ \\
 & Speculative & 16.99 & 1.11$\times$ & 16.86 & 1.88$\times$ \\
 & Lookahead & 25.14 & 1.64$\times$ & 13.77 & 1.54$\times$ \\
 & \name & \textbf{28.23} & \textbf{1.84}$\times$ & \textbf{24.03} & \textbf{2.68}$\times$ \\
\midrule
\multirow{4}{*}{CNN/DM} & Vanilla & 14.62 & 1.00 & 8.12 & 1.00 \\
 & Speculative & 17.82 & 1.22$\times$ & 12.77 & 1.57$\times$ \\
 & Lookahead & 18.77 & 1.28$\times$ & 9.47 & 1.17$\times$ \\
 & \name & \textbf{22.65} & \textbf{1.55}$\times$ & \textbf{14.67} & \textbf{1.81}$\times$ \\
\midrule
\multirow{4}{*}{WMT16} & Vanilla & 14.78 & 1.00$\times$ & 9.52 & 1.00$\times$ \\
 & Speculative & 17.48 & 1.18$\times$ & 14.72 & 1.55$\times$ \\
 & Lookahead & 17.98 & 1.22$\times$ & 14.65 & 1.54$\times$ \\
 & \name & \textbf{19.94} & \textbf{1.35}$\times$ & \textbf{19.27} & \textbf{2.02}$\times$ \\
\bottomrule
\end{tabular}}
\caption{The greedy decoding speed (token/s) and speedup ratio on GSM8K, CNN/DM and WMT16.}
\label{tab:greedy_text}
\end{table}
\egroup

We conduct experiments in both greedy decoding and random sampling scenarios. The experimental results are listed as follows.

\textbf{Greedy decoding}. As shown in 
Figure~\ref{fig:greedy_code}, in the greedy generation scenario, \name~outperforms vanilla greedy decoding, lookahead decoding, and speculative decoding under all backbone models and dataset configurations. \name~can achieve up to 61.2 token/s generation speed with the max generation length of 512, which achieves speedups of 3.9$\times$  compared to greedy decoding, 2.8$\times$ compared to speculative decoding, and 1.9$\times$ compared to lookahead decoding. 
Table~\ref{tab:greedy_text} shows that \name~can also get substantial speedups on typical natural language tasks, where lookahead decoding and speculative decoding can only achieve limited acceleration.

\textbf{Random sampling}.
We test \name~in the random sampling scenario using Llama-2-chat-70B/7B, considering that those models after SFT are more suitable for well understanding rich semantics distributed in human natural languages and generating diverse outputs.
For the sampling hyperparameters, the generation temperature is set among $0.5$ and $1.0$, and top-p is set to $0.8$.
Table~\ref{tab:random_text} shows that \name~can also be applied to random sampling, and the speedup over baseline methods are not much different from the observations in the greedy decoding scenario.


\bgroup
\small
\setlength{\tabcolsep}{4pt}
\begin{table}[t]
\center
\scalebox{0.76}{
\begin{tabular}{ll|c|cc}
\toprule
\multicolumn{1}{l}{\multirow{2}{*}{Task}}&\multicolumn{1}{l|}{\multirow{2}{*}{Algorithm}} & \multicolumn{1}{c|}{\makecell{Greedy}} & \multicolumn{2}{c}{\makecell{Random}} \\
& & (temp=0.0)  & temp=0.5 & temp=1.0 \\
\midrule
\multirow{4}{*}{GSM8k} & Vanilla & 8.96 & 8.97 & 8.96 \\
 & Speculative & 16.86 & 14.99 & 14.11 \\
 & Lookahead & 13.77 & 13.56 & 13.46 \\
 & \name & \textbf{24.03} & \textbf{22.04} & \textbf{19.27} \\
\midrule
\multirow{4}{*}{CNN/DM} & Vanilla & 8.12 & 8.83 & 8.30\\
 & Speculative & 12.77 & 13.43 & 13.75\\
 & Lookahead & 9.47 & 9.31 & 9.47 \\
 & \name & \textbf{14.67} & \textbf{14.97} & \textbf{14.23} \\
\midrule
\multirow{4}{*}{WMT16} & Vanilla & 9.52 & 9.75 & 9.75 \\
 & Speculative & 14.72 & 13.91 & 14.28\\
 & Lookahead & 14.65 & 12.05 & 11.91 \\
 & \name & \textbf{19.27} & \textbf{19.95} & \textbf{19.33} \\
\bottomrule
\end{tabular}}
\caption{The random sampling speed (token/s) compared with the greedy decoding speed (token/s), tested on Llama-2-chat 70B/7B. ``temp'' means the temperature used in the random sampling scenario.}
\label{tab:random_text}
\end{table}
\egroup

\subsection{Ablation Studies and Analyses}

In this section, to give a deeper insight into how \name~achieves higher generation speed, we conduct ablation studies and analyses to answer the following questions.

\bgroup
\small
\begin{table}[t]
\center
\scalebox{0.76}{
\begin{tabular}{l|c}
\toprule
Method & token/s \\
\midrule
Baseline & 21.46\\
\midrule
+ Accelerating drafting via phrases & 49.90 \\
+ Lengthening drafts via phrases & 55.92 \\
+ Generate phrases from verification & 58.18 \\
+ Reuse phrases from history context & 61.20 \\
\bottomrule
\end{tabular}}
\caption{The ablation studies of each component in \name~on HumanEval using Yi 34B/6B.}
\label{tab:ablation_components}
\end{table}
\egroup

\textbf{What are the effects of each component?} To demonstrate the specific speedup introduced by each mechanism, the ablation results are in Table \ref{tab:ablation_components}.
In the table, each component could bring a speedup gain, while accelerating drafting and lengthening the draft are the most effective ones.

\begin{figure}[t]
    \centering
    \includegraphics[width=0.9\linewidth]{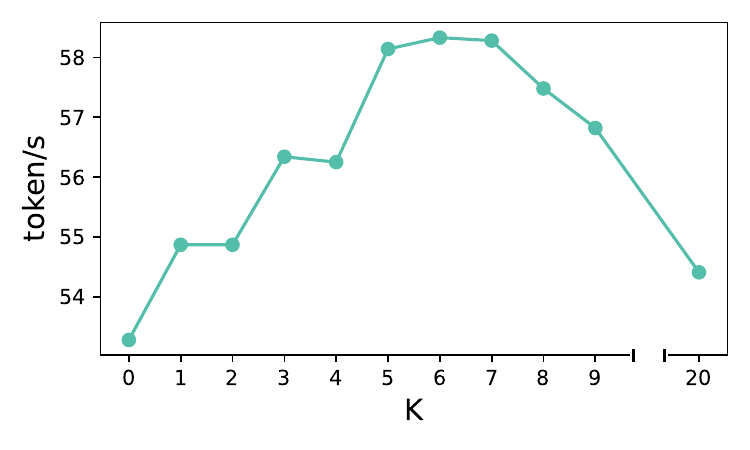}
    \caption{The effect of selecting $K$ phrases for Eq.~(\ref{eq:draft-process}), tested on HumanEval using Yi 34B/6B without reusing phrases from history contexts.}
    \label{fig:ablation_topk}
\end{figure}

\textbf{How many phrases are needed to lengthen the draft?}
Verifying too many phrases might slow down the verification in real-model inference scenarios. Here comes a trade-off between more phrases for possible speedup and slower verification.
In Figure~\ref{fig:ablation_topk}, there exists a best $K$ value, fewer or more phrases cause slower decoding speed.

\textbf{Other ablation experiments.}
The results of \name~on different hyperparameter and length settings are show in Appendix~\ref{sec:hyperparameters} and Appendix~\ref{sec:appendix_length}, respectively. The ablation of phrases reusing on different context locality are in Appendix~\ref{sec:appendix_context}.

\subsection{Comparing with Training-based Methods}

We compare our training-free \name~with the state-of-the-art training-based method Eagle~\cite{li2024eagle} on Spec-Bench~\cite{xia2024unlocking}. The main idea of Eagle is to specifically train a tiny draft model under their newly-designed model architecture and generate tree-style drafts for the target LLM. In order to accelerate Llama-2-chat-70B, it trains a 1B draft model by distilling from Llama-2-chat-70B.
The results in Table~\ref{tab:vs_training} show that, although Eagle trains a much smaller draft model with $1\over7$ of our scale to draft, it can only achieve the speed slightly faster than our training-free method. This is because the much smaller model significantly reduces the number of draft tokens accepted by the target model in each iteration, i.e., lossing draft effectiveness.
We look back at the training-based methods including Eagle, they pursue ultimate drafting speed at the expense of lossing draft accuracy.
We, on the other hand, optimize the drafting speed while keeping the accuracy unchanged.
Currently, Eagle cannot be integrated into \name~since the special model architecture of their draft model only supports autoregressive token-level drafting. We believe that our method can find a better balance between drafting speed and accuracy when combined with training, which will be our future work. 

\bgroup
\small
\small
\begin{table}[t]
\center
\scalebox{0.76}{
\begin{tabular}{l|cc|cc}
\toprule
\multicolumn{1}{l|}{\multirow{2}{*}{Spec-Bench}} & \multicolumn{2}{c|}{\makecell{\name}} & \multicolumn{2}{c}{\makecell{Eagle}} \\
& token/s & \#accept & token/s & \#accept \\
\midrule
MT-Bench & 24.23 & \textbf{5.16} & \textbf{28.20} & 3.52\\
Translation & 18.91 & \textbf{3.92} & \textbf{24.23} & 3.16\\
Summarization & 19.91 & \textbf{4.93} & \textbf{22.03} & 3.16 \\
QA & 21.63 & \textbf{4.67} & \textbf{24.83} & 3.23\\
Math Reasoning & 25.39 & \textbf{4.95} & \textbf{29.90} & 3.81 \\
RAG & 19.00 & \textbf{5.43} & \textbf{20.56} & 3.54 \\
\midrule
Average & 21.51 & \textbf{4.96} & \textbf{24.96} & 3.48 \\
\bottomrule
\end{tabular}}
\caption{The speed (token/s) comparing with training-based methods, tested on Llama-2-chat-70B. ``\#accept'' means the number of draft tokens accepted by the target model in each iteration (on average). \name~uses Llama-2-chat-7B as the draft model while Eagle uses its trained 1B model to draft.}
\label{tab:vs_training}
\end{table}
\egroup

\subsection{Comparing with Phrases-based Methods}

We compare \name~with other phrases-based method PLD~\cite{saxena2023prompt} and REST~\cite{he-etal-2024-rest} on Spec-Bench~\cite{xia2024unlocking}. These methods retrieve phrases from prompts or documents as drafts for the target model to verify. \name~outperforms these baselines by a large margin in Table~\ref{tab:vs_phrases}, which shows the effectiveness of using a draft model as an intermediary to filter away low-quality phrases before providing them to the target model.

\bgroup
\small
\small
\begin{table}[t]
\center
\scalebox{0.62}{
\begin{tabular}{l|cc|cc|cc}
\toprule
\multicolumn{1}{l|}{\multirow{2}{*}{Spec-Bench}} & \multicolumn{2}{c|}{\makecell{\name}} & \multicolumn{2}{c|}{\makecell{PLD}} & \multicolumn{2}{c}{\makecell{REST}} \\
& token/s & \#accept & token/s & \#accept & token/s & \#accept\\
\midrule
MT-Bench & \textbf{24.23} & \textbf{5.16} & 13.95 & 1.45 & 14.93 & 1.94\\
Translation & \textbf{18.91} & \textbf{3.92} & 12.99 & 1.41 & 13.02 & 1.57\\
Summarization & \textbf{19.91} & \textbf{4.93} & 16.98 & 1.97 & 12.60 & 1.69 \\
QA & \textbf{21.63} & \textbf{4.67} & 11.33 & 1.26 & 16.10 & 1.96\\
Math Reasoning & \textbf{25.39} & \textbf{4.95} & 15.86 & 1.75 & 13.07 & 1.60 \\
RAG & \textbf{19.00} & \textbf{5.43} & 15.51 & 1.64 & 13.28 & 1.91 \\
\midrule
Average & \textbf{21.51} & \textbf{4.96} & 14.44 & 1.51 & 13.83 & 1.83 \\
\bottomrule
\end{tabular}}
\caption{The speed (token/s) comparing with phrases-based methods, tested on Llama-2-chat-70B. ``\#accept'' means the number of draft tokens accepted by the target model in each iteration (on average).}
\label{tab:vs_phrases}
\end{table}
\egroup

\section{Related Work}

This section introduces existing efforts for efficient LLM inference, including efficient decoding, efficient implementation and model compression.

\subsection{Efficient Decoding}

To alleviate efficiency issues caused by autoregressive decoding, non-autoregressive decoding methods have been proposed.
Instead of generating tokens one by one, non-autoregressive methods generate multiple tokens in parallel at a time~\cite{wei2019imitation, guo2020jointly, ghazvininejad2019mask}. These non-autoregressive methods bring an improvement in inference efficiency and also significantly hurt model performance.
To this end, drafting-then-verifying methods have been proposed~\cite{blockwiseparallel,microsoftspec,googlespec,deepmindspec}. 
The drafting-then-verifying methods avoid LLMs from serial generation and instead use them to verify drafts in a non-autoregressive parallel manner, which do not reduce model performance and significantly accelerate inference.

The key to drafting-then-verifying methods is to generate drafts quickly and well.

\textbf{(1)}~
Methods such as speculative decoding~\cite{microsoftspec,googlespec,deepmindspec} use a model smaller than the target model to generate drafts. To further align the draft model with the target model, distillation techniques are applied~\cite{specinfer,distillspec}. Even much smaller models can be used as a draft to speed up the draft model, forming multi-staged speculative decoding~\cite{stagedspec,cascade}. Apart from using different models, Self-Speculative~\cite{selfspec} and PPD~\cite{predictpipeline} select part of the model layers as the draft model, while these methods requires extra training or pre-processing.

\textbf{(2)}~Some other efforts explore to use the target model itself to efficiently generate drafts. Blockwise~\cite{blockwiseparallel} and Medusa~\cite{medusa} train multiple extra output heads on top of LLMs and fine-tune heads to generate multiple draft tokens in parallel. 
Parallel Decoding~\cite{paralleldecoding} and PaSS~\cite{PaSS} add auxiliary input suffixes such as padding tokens or learnable padding tokens to generate draft output suffixes. Lookahead Decoding~\cite{fu2023lookahead} generates n-gram pools using Jacobi-iteration and uses those n-grams starting with the last generated token as drafts.

\textbf{(3)}~Other methods explore to retrieve phrases from previous prompts or existing documents, such as PLD~\cite{saxena2023prompt}, LLMA~\cite{yang2023inference}, and REST~\cite{he-etal-2024-rest}. These methods use the target model to directly verify phrases, incurring high failure costs on each draft trial. However, we are the first to use a draft model as an intermediary to filter away low-quality phrases before providing them to the target model.

Both the above methods are insufficiently efficient, so we first introduce a small model for drafting and then further accelerate the draft process of the small model by using phrases. We list various possibilities for the sources of phrases (from draft model, from verification, and from historical contexts) in our method and conduct ablation experiments for each source. This indicates the possibility that other phrase sources~\cite{saxena2023prompt,yang2023inference,he-etal-2024-rest} can be integrated into our framework to achieve further speedup.

Besides better generating drafts, traditional triangular attention masking can only verify one draft sentence using a complete forward pass. Tree-style verification~\cite{specinfer, medusa, fu2023lookahead} designs specific attention masking to verify multiple possible drafts at a time.

\subsection{Efficient Implementation}

The most direct solution to achieving efficient LLM inference is implementing LLMs efficiently to take advantage of hardware devices (such as GPUs).
FlashDecoding~\cite{flash-decoding} accelerates Transformer attention computation within LLMs by partitioning the decoding sequence into multiple small blocks and performing block-wise computation in fast GPU SRAM instead of GPU HBM. 
PageAttention~\cite{vllm} using paged virtual memory management to organize the decoding cache during the decoding process, thereby effectively utilizing GPU memory bandwidth to reduce the memory access overhead of inference.
Tensor Parallelism~\cite{megatron} accelerates inference by sharding matrices into distributed GPUs and performing matrix multiplications in a distributed manner.
Some efforts implement LLMs by optimizing underlying operators~\cite{fasttransformer,wang2021lightseq,tensorrtllm} and achieve promising results.
Note that these efficient implementations of LLMs and \name~are orthogonal. Combining efficient implementation methods and \name~can achieve more significant inference acceleration.

\subsection{Model Compression}

Model compression methods are proposed to reduce the number of operations needed for model execution. 
Structure pruning~\cite{Fan2020Reducing,wang-etal-2020-structured,ZHANG202136,xia-etal-2022-structured} and unstructured pruning~\cite{Han2015LearningBW,NEURIPS2020_b6af2c97,xu-etal-2021-rethinking} elimate non-essential parameters. 
Quantization~\cite{zafrir2019q8bert,frantar2023gptq,awq,squeezellm,stock2021training} methods quantize parameters into low-bit representations. 
Distillation~\cite{hinton2015distilling,sun-etal-2019-patient,jiao-etal-2020-tinybert,liu-etal-2022-multi-granularity, park-etal-2021-distilling} methods are used to help align those compressed models to their original versions to maintain model performance. 
Early-exiting~\cite{depth-adaptive,FREE} methods end the inference process when the output result in shallow layers reaches the confidence threshold.
MoEfication~\cite{zhang-etal-2022-moefication,song2023powerinfer} turns a dense model into a sparse activated model.
Model compression and \name~are also orthogonal and can be integrated for further acceleration.

\section{Conclusion}

In this paper, we propose a practical algorithm \name~to improve drafting efficiency for speculative decoding. We generate phrases in a heuristic manner and use phrases to help draft models accelerate drafting and lengthen drafts.
Our experiments verify that, compared to typical baselines vanilla speculative decoding and lookahead decoding, \name~respectively achieves speedups of 2.8$\times$ and 1.9$\times$ and does not affect the generation quality at all. Our method is completely training-free, to make it easier to adopt for the users of speculative decoding. And the training-based version of \name~towards extreme acceleration will be our next work.

\section*{Limitations}
We only focus on decoder-only model structure in the experiment.
However, \name~is an extendable framework, which can be applied to various model structures. Our method currently focuses on optimization in training-free scenarios. If the consistency of large and small models is improved through training in the future, our method can achieve higher acceleration.
Both efficient implementation and model compression are orthogonal to our method, and can be combined for further speedup.

We only focus on the single query scenario. The application of speculative sampling in the batched inference scenario is not within the scope of this paper and can refer to~\cite{liu2024optimizing,qian2024bass,chen2024magicdec}.

\section*{Ethical Considerations}

Our method has no potential risk since we are training-free and does not affect the generation results at all. 

\section*{Acknowledgments}
This work was supported by the National Key R\&D Program of China (2022ZD0160501), Institute Guo Qiang at Tsinghua University. Yuxiang Huang is supported by Tsinghua University Initiative Scientific Research Program (Student Academic Research Advancement Program).
We acknowledge valuable discussions with Yuhui Li, the author of EAGLE.

\bibliography{custom}

\clearpage
\appendix

\section{\name~does not cause any performance degradation}

\bgroup
\setlength{\tabcolsep}{4pt}
\begin{table}[ht]
\small
\center
\scalebox{0.85}{
\begin{tabular}{ll|ccc}
\toprule
\multicolumn{2}{c|}{Model} & \makecell{Yi\\34B/6B} & \makecell{DeepSeek\\33b/6.7B} & \makecell{CodeLlama\\34B/7B}\\
\midrule
\multirow{4}{*}{HumanEval} & Greedy & 18.3 & 64.6 & 36.0\\
& Speculative & 18.9 & 64.6 & 36.0\\
& Lookahead & 18.9 & 64.6 & 36.0\\
& \name & 18.9 & 64.6 & 36.0\\
\midrule
\multirow{4}{*}{MBPP} & Greedy & 37.8 & 70.0 & 46.6\\
& Speculative & 38.9 & 71.1 & 46.6\\
& Lookahead & 38.9 & 71.1 & 46.6\\
& \name & 38.9 & 71.1 & 46.6\\
\bottomrule
\end{tabular}}
\caption{Task performance of HumanEval and MBPP of Greedy, Speculative, Lookahead and \name~using greedy generation. We report Accuracy(\%) of HumanEval and MBPP.}
\label{tab:performance -- codegen}
\end{table}
\egroup

\bgroup
\begin{table}[ht]
\small
\center
\scalebox{0.85}{
\begin{tabular}{ll|cc}
\toprule
\multicolumn{2}{c|}{Model} & \makecell{Yi\\34B/6B} & \makecell{Llama\\70B/7B} \\
\midrule
\multirow{4}{*}{GSM8K} & Greedy & 61.0 & 66.0 \\
& Speculative & 61.0 & 66.0 \\
& Lookahead & 61.0 & 66.0 \\
& \name & 61.0 & 65.0 \\
\midrule
\multirow{4}{*}{CNN/DM} & Greedy & 35.2 & 37.1\\
& Speculative & 35.1 & 37.2\\
& Lookahead & 35.3 & 37.1\\
& \name & 35.3 & 37.1\\
\midrule
\multirow{4}{*}{WMT16} & Greedy & 21.7 & 31.1\\
& Speculative & 21.9 & 31.1\\
& Lookahead & 22.0 & 31.1\\
& \name & 21.1 & 31.1\\
\bottomrule
\end{tabular}}
\caption{Task performance of GSM8K, CNN/DM and WMT16 of Greedy, Speculative, Lookahead and \name~using greedy generation. We report Accuracy of GSM8K(\%), Rouge1 score(\%) of CNN/DM and BLEU score(\%) of WMT16.}
\label{tab:performance -- text}
\end{table}
\egroup

\bgroup
\begin{table}[ht]

\small
\center
\scalebox{0.85}{
\begin{tabular}{ll|c}
\toprule
\multicolumn{2}{c|}{Model} & \makecell{Yi\\34B/6B} \\
\midrule
\multirow{4}{*}{HumanEval} & Random Sampling & 15.9\\
& Speculative & 18.9\\
& Lookahead & 18.9\\
& \name & 17.7\\
\midrule
\multirow{4}{*}{MBPP} & Random Sampling & 
 36.7\\
& Speculative & 33.3\\
& Lookahead & 35.6\\
& \name & 36.7\\
\midrule
\multirow{4}{*}{GSM8K} & Random Sampling & 
 67.0\\
& Speculative & 70.0\\
& Lookahead & 69.0\\
& \name & 69.0\\
\midrule
\multirow{4}{*}{CNN/DM} & Random Sampling & 34.6\\
& Speculative & 34.0\\
& Lookahead & 34.6\\
& \name & 35.6\\
\midrule
\multirow{4}{*}{WMT16} & Random Sampling & 
 20.6\\
& Speculative & 21.2\\
& Lookahead & 22.7\\
& \name & 25.0\\
\bottomrule
\end{tabular}}
\caption{Task performance using random sampling. We report accuracy for HumanEval, MBPP and GSM8K, Rouge 1 score for CNN/DM and BLEU score for WMT16.}
\label{tab:performance -- random}
\end{table}
\egroup

Table \ref{tab:performance -- codegen}, \ref{tab:performance -- text} and~\ref{tab:performance -- random}~show the task performance of the experiments shown in Figure~\ref{fig:greedy_code} and Table~\ref{tab:greedy_text}. Theoreotically, Speculative Decoding, Lookahead Decoding and \name~generates the same token sequences compared with vanilla~Decoding. We observe no overall performance degradation of \name~ from Table~\ref{tab:performance -- codegen}, \ref{tab:performance -- text} and~\ref{tab:performance -- random}. The small difference in task performance among vanilla, Speculative, Lookahead and \name~ is caused by floating point error, since the calculations are not fully the same between the methods (especially the calculation order of attention). The experimental results show that the calculation error would not cause notable performance degradation.

\newpage 
\section{Hyperparameters}
\label{sec:hyperparameters}

$W$ is a hyperparameter used in Lookahead Decoding to generate phrases. Larger $W$ indicates more phrases are generated in each forward of our draft model in \name, but may also be more time-consuming.

To investigate hyperparameters sensitivity, we conduct a grid search on $\gamma, W$ and $\beta$ on GSM8K with Yi-34B/6B. We report token/s of each setting.

\begin{table}[h]
\center
\scalebox{0.55}{
\begin{tabular}{l|ccc|ccc|ccc}
\toprule
& \multicolumn{3}{c|}{$W=14$} & \multicolumn{3}{c|}{$W=15$} & \multicolumn{3}{c}{$W=16$} \\
& $\beta=5$ & $\beta=6$ & $\beta=7$ & $\beta=5$ & $\beta=6$ & $\beta=7$ & $\beta=5$ & $\beta=6$ & $\beta=7$ \\
\midrule
$\gamma=3$ & 28.17 & 28.07 & 27.30 & 27.97 & 28.23 & 27.38 & 28.34 & 27.64 & 27.21 \\
$\gamma=4$ & 28.52 & 28.42 & 27.91 & 29.06 & 28.77 & 28.23 & 29.00 & 28.68 & 27.64\\
$\gamma=5$ & 28.13 & 29.32 & 28.20 & 28.31 & 28.96 & 27.99 & 27.94 & 29.03 & 27.90\\
\midrule
\multicolumn{5}{c}{token/s std: 0.57} & \multicolumn{5}{c}{speedup std: 0.04}\\
\bottomrule
\end{tabular}}
\caption{The results of grid search where $\gamma\in [3,5], W\in [14, 16], \beta\in [5, 7]$. ``std'' represents the standard deviation.}
\label{tab:grid search}
\end{table}

Table \ref{tab:grid search} shows that \name~is stable to hyperparameters. This provides an opportunity to provide a general recipe or conduct heuristic search instead of grid search to find a good combination of hyperparameters within a short time.

\subsection{Hyperparameters Recipe}

Here, we offer a general recipe for hyperparameter tuning in Table~\ref{tab:recipe}. We distinguish downstream tasks into two types: (1) \textbf{HH}: High homogeneity between draft and target model, e.g. code generation; (2) \textbf{LH}: Low homogeneity between draft and target model, e.g. machine translation. 

\bgroup
\begin{table}[h]
\small
\center
\scalebox{0.85}{
\begin{tabular}{l|cccc}
\toprule
Tasks & $\gamma$ & $W$ & $\beta$ & $K$\\
\midrule
HH & $7\sim 14$ & $15\sim 20$ & $5\sim 7$ & $3\sim 5$ \\
LH & $2\sim 6$ & $15\sim 20$ & $5\sim 7$ & $3\sim 5$ \\
\bottomrule
\end{tabular}}
\caption{An empirical recipe for hyperparameters tuning.}
\label{tab:recipe}
\end{table}
\egroup

\subsection{Hyperparameters Tuning}
\label{sec:hyper-tuning}
According to the stability of hyperparameters, we conduct a heuristic search to find a good combination. The algorithm of the heuristic search is described in Algorithm~\ref{alg:heur}.

\begin{algorithm}[h]

\small
\DontPrintSemicolon
\caption{Heuristic Search of Hyperparameters}
\label{alg:heur}
\KwIn{Task type $T$, Dataset $D$, Objective function $f$.}
\KwOut{Hyperparameters $\gamma, W, \beta, K$}
$K_0$ $\coloneqq$ $3$ \tcp{$K$ is always $3$ in our experiments}
$\hat{W} \sim [15, 20]$\;
$\hat{\beta} \sim [5, 7]$\;
\eIf{$T=$ HH}{
    $\hat{\gamma} \sim [7, 14]$\;
}{
    $\hat{\gamma} \sim [2, 6]$\;
}
$\gamma_0\coloneqq \arg\min_{\gamma}\text{ClockTime}(f(D; \hat{\gamma}, \hat{W}, \hat{\beta}, K_0))$\;
$W_0\coloneqq \arg\min_{W}\text{ClockTime}(f(D; \gamma_0, \hat{W}, \hat{\beta}, K_0))$\;
$\beta_0\coloneqq \arg\min_{\beta}\text{ClockTime}(f(D; \gamma_0, W_0, \hat{\beta}, K_0))$\;
\Return $\gamma_0, W_0, \beta_0, K_0$\;

\end{algorithm}

\subsection{Hyperparameters Used in Our Experiment}

The hyperparameters of our experiments are shown in Table~\ref{tab:hyperparameters -- main experiments}. In Table \ref{tab:ablation_components}, we use $\gamma = 12, W=20, \beta=8$ and $K=3$. In Figure \ref{fig:ablation_topk}, we use $\gamma = 12, W=20, \beta=8$. In Table~\ref{tab:locality}, we use $\gamma = 6, W=18, \beta=6$ and $K=3$.
In Table~\ref{tab:block efficiency}, we follow the hyperparameters in Table~\ref{tab:hyperparameters -- main experiments}.

\bgroup
\setlength{\tabcolsep}{4pt}
\begin{table}[ht]
\small
\center
\scalebox{0.74}{
\begin{tabular}{ll|c|cc|cccc}
\toprule
\multirow{2}{*}{Task} & \multirow{2}{*}{Model} & Speculative & \multicolumn{2}{c|}{Lookahead} & \multicolumn{4}{c}{\name} \\
& & $\gamma$ & $W$ & $\beta$ & $\gamma$ & $W$ & $\beta$ & $K$\\
\midrule
\multirow{3}{*}{HumanEval} & Yi-34B/6B & 12 & 16 & 7 & 12 & 20 & 8 & 3\\
& DeepSeek-33b/6.7B & 3 & 16 & 7 & 11 & 20 & 8 & 3\\
& CodeLlama-34B/7B & 5 & 17 & 5 & 10 & 20 & 7 & 3\\
\midrule
\multirow{3}{*}{MBPP} & Yi-34B/6B & 6 & 16 & 6 & 6 & 18 & 6 & 3\\
& DeepSeek-33b/6.7B & 5 & 15 & 7 & 7 & 15 & 6 & 3\\
& CodeLlama-34B/7B & 4 & 16 & 6 & 8 & 16 & 7 & 3\\
\midrule
\multirow{2}{*}{GSM8K} & Yi-34B/6B & 7 & 15 & 6 & 4 & 15 & 7 & 3\\
& Llama-70B/7B & 7 & 15 & 6 & 6 & 17 & 7 & 3\\
\midrule
\multirow{2}{*}{CNN/DM} & Yi-34B/6B & 5 & 15 & 6 & 10 & 15 & 6 & 3\\
& Llama-70B/7B & 4 & 16 & 5 & 4 & 13 & 6 & 3\\
\midrule
\multirow{2}{*}{WMT16} & Yi-34B/6B & 2 & 15 & 6 & 5 & 16 & 6 & 3\\
& Llama-70B/7B & 5 & 18 & 7 & 4 & 13 & 6 & 3\\
\bottomrule
\end{tabular}}
\caption{Hyperparameters of experiments in Figure~\ref{fig:greedy_code} and Table \ref{tab:greedy_text}.}
\label{tab:hyperparameters -- main experiments}
\end{table}
\egroup

\section{Block Efficiency}
In order to show the potential of further systematic optimization, we follow~\citet{specinfer} to test the Block Efficiency, with the hypothesis that the drafting time can be fully ignored. Block efficiency is the theoretical upper bound of the acceleration ratio, without considering any resource constraints.

\begin{definition}[Block Efficiency] Define Block Efficiency $\eta$ as
    $$\eta = \frac{\text{\# Generated tokens}}{\text{\# Target model calls}}$$
\end{definition}

We measure Block Efficiency with Yi-34B/6B on all 5 datasets following the settings of Figure \ref{fig:greedy_code} and Table \ref{tab:greedy_text}. Results are in Table~\ref{tab:block efficiency}.

\bgroup
\small
\setlength{\tabcolsep}{4pt}
\begin{table}[h]
\center
\scalebox{0.85}{
\begin{tabular}{l|c|c|c}
\toprule
& Speculative & Lookahead & \name \\
\midrule
HumanEval & \underline{11.16} & 3.08 & \textbf{13.12} \\
MBPP & \underline{6.14} & 2.71 & \textbf{7.43} \\
GSM8K & \textbf{5.23} & 2.30 & \underline{4.65} \\
CNN/DM & \underline{4.71} & 1.70 & \textbf{7.46} \\
WMT16 & \underline{2.41} & 1.37 & \textbf{4.05}\\
\bottomrule
\end{tabular}}
\caption{The Block Efficiency $\eta$ of Yi-34B/6B on HumanEval, MBPP, GSM8K, CNN/DM and WMT16. The method with largest $\eta$ is highlighted with the \textbf{bold} font, and the second largest result is with the \underline{underlined} font.}
\label{tab:block efficiency}
\end{table}
\egroup

From Tabel \ref{tab:block efficiency}, \name~has relatively high Block Efficiency in most scenarios, showing that it has a larger ideal speedup ratio that may be achieved by future algorithms or systematic implementation.

\section{Comparing with other draft-accelerated methods}

Cascade Speculative Decoding~\cite{cascade} is a typical method that accelerates the draft model in speculative decoding by further using an even smaller model to draft for the draft model. The comparison between \name~and Cascade is described in Table~\ref{tab:cascade}. Results show that \name~outperforms Cascade, showing that the way we accelerate the draft is more efficient.

Cascade performs slower than the original speculative decoding.
One possible reason is that the tiny models which are utilized to accelerate the Llama-2-7B model, i.e. Llama-160M~\cite{specinfer} and TinyLlama-1.1B~\cite{zhang2024tinyllama}, are not officially trained by Meta. This has led to discrepancies in the model's output, thereby slowing down the drafting process.

\bgroup
\small
\setlength{\tabcolsep}{4pt}
\begin{table}[h]
\center
\scalebox{0.8}{
\begin{tabular}{c|r|c}
\toprule
Method & Model Series & token/s \\
\midrule
Speculative & 7B$\to$70B & 16.86 \\
\midrule
Cascade & 160M$\to$7B$\to$70B & 7.90 \\
Cascade & 1.1B$\to$7B$\to$70B &  9.22 \\
\midrule
\name & 7B$\to$70B & \textbf{24.03} \\
\bottomrule
\end{tabular}}
\caption{Comparing with various cascade versions of speculative decoding, tested on GSM8K.}
\label{tab:cascade}
\end{table}
\egroup

\section{\name~under various generation lengths}
\label{sec:appendix_length}

To investigate the speedup ratio under different generation length limits, we conduct the following experiment with generation length limits 64, 128, 256, and 512. This experiment is performed on GSM8K with Yi-34B/6B, with other hyperparameters aligned with Table~\ref{tab:hyperparameters -- main experiments}.

\bgroup
\setlength{\tabcolsep}{4pt}
\begin{table}[h]
\small
\center
\scalebox{0.78}{
\begin{tabular}{ll|cccc}
\toprule
& & Greedy & Speculative & Lookahead & \name\\
\midrule
\multirow{2}{*}{$l=64$} & token/s & 14.31 & 20.18 & 20.20 & \textbf{30.99}\\
& speedup & 1.00 & 1.41 & 1.41 & \textbf{2.17}\\
\midrule
\multirow{2}{*}{$l=128$} & token/s & 14.57 & 21.90 & 22.68 & \textbf{35.32}\\
& speedup & 1.00 & 1.50 & 1.56 & \textbf{2.42}\\
\midrule
\multirow{2}{*}{$l=256$} & token/s & 14.41 & 22.12 & 24.58 & \textbf{38.36}\\
& speedup & 1.00 & 1.54 & 1.71 & \textbf{2.66}\\
\midrule
\multirow{2}{*}{$l=512$} & token/s & 15.35 & 21.45 & 25.14 & \textbf{40.94}\\
& speedup & 1.00 & 1.40 & 1.64 & \textbf{2.67}\\
\bottomrule
\end{tabular}}
\caption{Speedup with multiple generation lengths. ``$l$'' represents generation length limit.}
\label{tab:multiple-length -- speed}
\end{table}
\egroup

\bgroup
\setlength{\tabcolsep}{4pt}
\begin{table}[h]
\small
\center
\scalebox{0.78}{
\begin{tabular}{l|cccc}
\toprule
 & Greedy & Speculative & Lookahead & \name\\
\midrule
$l=64$ & 44.51 & 44.51 & 44.51 & 44.51\\
$l=128$ & 60.98 & 60.98 & 60.98 & 60.98\\
$l=256$ & 64.02 & 64.02 & 63.41 & 63.41\\
$l=512$ & 64.63 & 64.63 & 64.02 & 64.63\\
\bottomrule
\end{tabular}}
\caption{Task performance (accuracy) with multiple generation lengths. ``$l$'' represents generation length limit.}
\label{tab:multiple-length -- performance}
\end{table}
\egroup

From Table~\ref{tab:multiple-length -- speed}, \name~outperforms autoregressive decoding, Speculative decoding, and Lookahead decoding under various lengths. In Table~\ref{tab:multiple-length -- performance}, short generation length limit ($l=64, 128$) results in answer truncation, which leads to significant lower task performance. Close task performances and speedup ratios are achieved in $l=256, 512$, showing that all algorithms halt immediately after generating ``$<$EOS$>$''.

Two conclusions can be obtained. First, the length limits used in our experiments are reasonable, as shorter length limits result in ill-generated answers. Second, phrase repetition in the generation is not the reason why \name~achieves such a high speedup. If so, we would be able to observe lower task performance and much higher speedup from $l=256$ to $l=512$.

\section{Context Locality}
\label{sec:appendix_context}

\begin{figure}[t]
    \centering
    \includegraphics[width=0.8\linewidth]{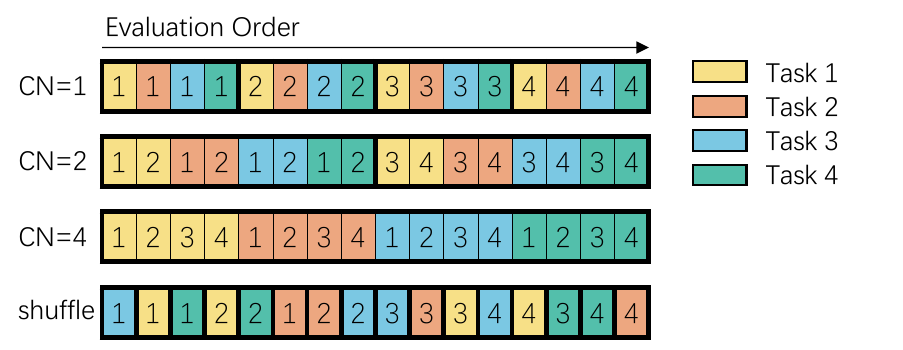}
    \caption{The ablation studies on context locality. In our experiments, the task 1,2,3,4 are MBPP, GSM8K, CNN/DM, and WMT16, respectively.}
    \label{fig:locality}
\end{figure}

We further study how context locality accelerate generation through the reused phases from history.

Context locality remains very high when continually generating within one task, but may vary across different datasets. For example, the phrases pool should contain code pieces when running on code generation datasets such as HumanEval and MBPP, but it might contain natural language pieces when generating text. It remains a question that whether one type of language pieces could accelerate another type of language's generation.
We conduct the following experiments to further investigate to what extent and how context locality can accelerate generation through reusing phrases from history.

We select four datasets: MBPP, GSM8K, CNN/DM, and WMT16, corresponding to code generation, arithmetic reasoning, document summarization, and machine translation, and sample 20 entries from each dataset to organize a new dataset containing different domains. We evaluate \name~with multiple evaluation order, from which we could change context locality. We define consecutive number CN as how many entries from the same dataset are tested consecutively, as shown in Figure~\ref{fig:locality}. In the ``shuffle'' configuration, we randomize the order of data entries. Higher CN indicates a better locality, and the ``shuffle'' configuration should have the worst locality. We then measure the generation speed in different CN configurations or random shuffling.

\begin{table}[h]
\center
\scalebox{0.5}{
\begin{tabular}{l|cc|ccccc}
\toprule
Setting & shuffle & CN=20 & shuffle & CN=20 & CN=10 & CN=4 & CN = 1\\
\midrule
Phrases Reusing & off & off & on & on & on & on & on\\
\midrule
token/s & 32.68 & 32.53 & 35.39 & 36.00 & 35.32 & 34.83 & 35.36\\
\bottomrule
\end{tabular}}
\caption{The results of ablation studies on context locality. ``CN'' stands for consecutive number.}
\label{tab:locality}
\end{table}

Table \ref{tab:locality} shows that context locality indeed affects the effectiveness of phrases reusing. With phrases reusing, CN=20 leads to faster decoding compared to the shuffle setting.
However, the effect caused by context locality is smaller than whether to turn on the phrases reusing. The speed of lower CN is similar to the shuffle setting when applying phrases reusing, but both are approximately 3 token/s faster compared to the cold start setting, proving that the phrases pool is still effective across multiple tasks. 

\end{document}